\documentclass[final]{cvpr}

\usepackage{times}
\usepackage{epsfig}
\usepackage{graphicx}
\usepackage{amsmath}
\usepackage{amssymb}
\usepackage{float}
\usepackage{multirow}
\usepackage{mathtools}
\usepackage{hyperref}
\hypersetup{
    colorlinks=true,
    linkcolor=blue,
    filecolor=blue,      
    urlcolor=blue,
    citecolor=cyan,
}
\def\cvprPaperID{3271} % *** Enter the CVPR Paper ID here
\def\confYear{CVPR 2021}

\begin{document}

%%%%%%%%% TITLE
\title{SOE-Net: A Self-Attention and Orientation Encoding Network \\for Point Cloud based Place Recognition}

% \author{Yan Xia \\ 
% Technical University of Munich\\
% % Institution1\\
% {\tt\small yan.xia@tum.de}
% \and
\author{
	\small
	\begin{tabular}{c c c c c c c }                                                          
		Yan Xia$^1$ &
		Yusheng Xu$^{1 }$ &
		Shuang Li$^2$ &
		Rui Wang$^{1 }$ &
		Juan Du$^1$ &
		Daniel Cremers$^1$ &
		Uwe Stilla$^1$ \\                                        
		\multicolumn{7}{c}{$^1$Technical University of Munich $^2$Beijing Institute of Technology} \\                                                
		\multicolumn{7}{c}{\{yan.xia, yusheng.xu, stilla\}@tum.de, \{wangr, duj, cremers\}@in.tum.de, shuangli@bit.edu.cn} 
% 		\multicolumn{7}{c}{\{wangr, duj, cremers\}@in.tum.de} \\
% 		\multicolumn{7}{c}{shuangli@bit.edu.cn} \\
% 		\multicolumn{7}{c}{Note: the author list is just a placeholder, and order needs to be discussed.} \\
	\end{tabular}                                                                       
}
\maketitle

% \renewcommand{\thefootnote}{\fnsymbol{footnote}} 
% \footnotetext[2]{Corresponding author.} 
\pagestyle{empty}  % no page number for the second and the later pages
\thispagestyle{empty} % no page number for the first page

%%%%%%%%% ABSTRACT
\begin{abstract}
We tackle the problem of place recognition from point cloud data and introduce a self-attention and orientation encoding network (SOE-Net) that fully explores the relationship between points and incorporates long-range context into point-wise local descriptors.
Local information of each point from eight orientations is captured in a PointOE module, whereas long-range feature dependencies among local descriptors are captured with a self-attention unit. 
Moreover, we propose a novel loss function called Hard Positive Hard Negative quadruplet loss (HPHN quadruplet), that achieves better performance than the commonly used metric learning loss.
% Experiments on various benchmark datasets demonstrate promising performance of the proposed network. 
% \textcolor{red}{It significantly outperforms the current state-of-the-art approaches on the Oxford RobotCar dataset.} 
Experiments on various benchmark datasets demonstrate superior performance of the proposed network over the current state-of-the-art approaches. 
Our code is released publicly at \url{https://github.com/Yan-Xia/SOE-Net}.
\end{abstract}

%%%%%%%%% BODY TEXT
\section{Introduction}

\begin{figure}[ht!]
\centering
\includegraphics[width=8.3cm]{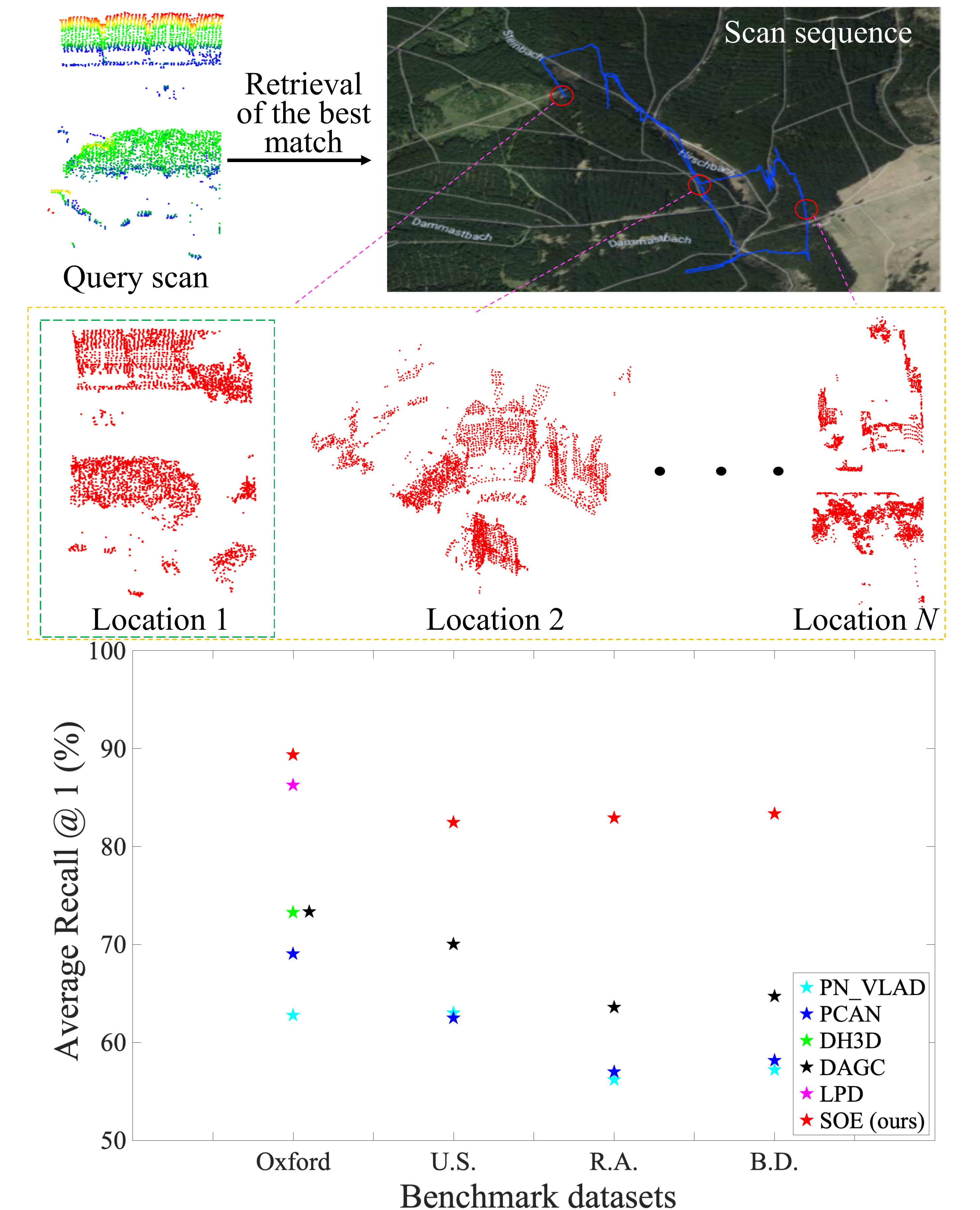}    
\caption{(Top) Place recognition from 3D point clouds: the street scene of a route map (shown in blue line) is denoted by a set of real-scan point clouds tagged with UTM coordinates. Given a query scan, we retrieve the closest match in this map (shown in the green box) to get the location of the query scan. (Bottom) Comparing the average recall at top 1 retrieval, SOE-Net significantly outperforms all published methods on various datasets.}
% \protect\footnotemark[1].} 
\label{fig: cover} 
\end{figure}

Place recognition and scene localization in large-scale and complex environments is a fundamental challenge with applications ranging from autonomous driving \cite{hane20173d, hee2013motion, liu2017robust} and robot navigation \cite{fu2018texture, mur2015orb} to augmented reality \cite{liu2016robust}. 
Given a query image or a LiDAR scan, the aim is to recover the closest match and its location by traversing a pre-built database. 
In the past decade, a variety of image-retrieval based solutions have shown promising performance \cite{li2010location, liu2017efficient,liu2019stochastic}. 
However, the performance of image-based methods often degrades when facing drastic variations in illumination and appearance caused by weather and seasonal changes \cite{angelina2018pointnetvlad}. 
As a possible remedy, 3D point clouds acquired from LiDAR offer accurate and detailed 3D information that is inherently invariant to illumination changes. 
As a consequence, place recognition from point cloud data is becoming an increasingly attractive research topic. Fig.~\ref{fig: cover}~(Top) shows a typical pipeline for point cloud based place recognition: One first constructs a database with LiDAR scans tagged with UTM coordinates acquired from GPS/INS readings. Given a query LiDAR scan, we then retrieve the closest match and its corresponding location from the database. 

The main challenge of point cloud based place recognition lies in how to find a robust and discriminative global descriptor for a local scene point cloud. While there exist abundant works on learning image descriptors, learning from point cloud data is far less developed.
To date, only a few networks have been proposed for point cloud based place recognition in large-scale scenarios. 
PointNetVLAD \cite{angelina2018pointnetvlad} is a pioneering work, which first extracts the local features from 3D point clouds using PointNet \cite{qi2017pointnet} and then fuses them into global descriptors using the NetVLAD \cite{arandjelovic2016netvlad} layer. 
PCAN \cite{zhang2019pcan} adapts the PointNet++ \cite{qi2017pointnet++} architecture to generate an attention map which re-weights each point during the local descriptors aggregation stage. 
% However, both methods neglect the geometric-relationship among points, therefore they cannot capture local features very well. 
Both methods use PointNet \cite{qi2017pointnet} to extract local descriptors, which ignores the geometric relationship among points. 
As of late, the authors of DH3D \cite{du2020dh3d} and DAGC \cite{sun2020dagc} noticed this shortcoming and designed  advanced local feature description networks. 
While this results in better local descriptors, both approaches simply aggregate these local descriptors to a global descriptor using the PCAN or PointNetVLAD fusion architecture, without considering the long-range dependencies of different features.

Similar to these previous studies \cite{du2020dh3d,sun2020dagc}, we also notice the importance of better utilizing the neighborhood context of each point when extracting its geometric representation.
To tackle this problem, we adopt a point orientation encoding (PointOE) module to encode the neighborhood information of various orientations. The orientation-encoding unit is integrated with PointNet \cite{qi2017pointnet}, taking its advantages in feature representation learning provided by the multi-layer perceptrons.
Another observation is, when aggregated into a global descriptor, different local descriptors should tactically contribute unevenly.
To achieve this, we develop a self-attention unit to introduce long-range contextual dependencies, encoding the spatial relationships of the local descriptors for weighting.
Combining the principals above, we propose a novel network named SOE-Net (Self-attention and Orientation Encoding Network). It is an end-to-end architecture that explores the relationship among the raw 3D points and the different importance of local descriptors for large-scale point cloud based retrieval. 
Specifically, SOE-Net combines local descriptor extraction and aggregation, which enables one-stage training to generate a discriminative and compact global descriptor from a given 3D point cloud. 
Additionally, we propose a novel ``Hard Positive Hard Negative quadruplet'' (HPHN quadruplet) loss, which addresses some of the limitations of the widely used lasy quadruplet loss. 
To summarize, main contributions of this work include:
\begin{itemize}
    \item We propose a novel point orientation encoding (PointOE) module to effectively extract local descriptors from a given point cloud, considering the relationship between each point and its neighboring points. 
    We further design a self-attention unit to differentiate the importance of different local descriptors to a global descriptor. 
    \item We present a new loss function termed HPHN quadruplet loss that is more effective for large-scale point cloud based retrieval.
    Comparing with previous loss functions, it can achieve more versatile global descriptors by relying on the maximum distance of positive pairs and the minimum distance of negative pairs.   
    \item We conduct experiments on four benchmark datasets, including Oxford RobotCar \cite{maddern20171} and three in-house datasets to demonstrate the superiority of SOE-Net over other state-of-the-art methods. 
    Notably, the performance on Oxford RobotCar reaches a recall of 89.37\% at top 1 retrieval.
\end{itemize}

% We notice that the local descriptors extracted by previously proposed networks \cite{zhang2019pcan, angelina2018pointnetvlad} are all point-wise global features. They neglect 

\section{Related work} \label{sec: related work}
Usually, the implementation of place recognition based on 3D point cloud retrieval is converted to a 3D feature matching problem, in which the 3D descriptor has substantial impact on the matching performance. Numerous methods for extracting 3D descriptors from point clouds have been developed, which can be roughly grouped into two categories: local descriptors and global descriptors.

{\bf 3D local descriptors.} Encoding robust local geometric information has long been a core challenge in robotics and 3D vision, with various attempts made. 
For example, extracting local structural information as histograms is representative. 
Spin Image (SI) \cite{johnson1999using} deploys the spin image representation to match 3D points. 
Geometry Histogram \cite{frome2004recognizing} proposes a novel regional shape context descriptor to improve the 3D object recognition rate on noisy data. 
Point Feature Histograms (PFH) \cite{rusu2008aligning} and Fast Point Feature Histograms (FPFH) \cite{rusu2009fast} seek to calculate the angular features and surface normals to represent the relationship between a 3D point and its neighbors. 
However, these handcrafted descriptors are unsuitable for large-scale scenarios due to the computational cost, at the mean time which are also sensitive to noisy and incomplete data acquired by sensors. 
Recently, learning-based methods for 3D local descriptor extraction have gained significant developments boosted by large-scale 3D datasets. 
3DMatch \cite{zeng20173dmatch} converts 3D points to voxels and then uses a 3D convolution network for segment matching. 
PPFNet \cite{deng2018ppfnet} and PPF-FoldNet \cite{deng2018ppf} directly use the raw 3D points as input and learn point pair features from points and normals of local patches. 
Fully Convolutional Geometric Features (FCGF) \cite{choy2019fully} proposes a compact geometric feature computed by a 3D fully-convolutional network. 
3DFeatNet \cite{yew20183dfeat} designs a weakly supervised network to learn both the 3D feature detector and descriptor. 
Furthermore, ASLFeat \cite{Luo_2020_CVPR} focuses on mitigating limitations in the joint learning of 3D feature detectors and descriptors. 
3DSmoothNet \cite{gojcic2019perfect} and DeepVCP \cite{lu2019deepvcp} learn compact and rotation invariant 3D descriptors relying on 3D CNNs. RSKDD-Net~\cite{rskdd} introduces random sampling concept to efficiently learn keypoint detector and descriptor.
Some methods explore to compress the dimensions of handcrafted 3D local descriptors utilizing deep learning, such as Compact Geometric Features (CGF) \cite{khoury2017learning} and LORAX \cite{elbaz20173d}. %However, these local descriptors are hard to be applied to extract global features due to the data increasing \cite{zhang2019pcan}.

{\bf 3D global descriptors.} Different from 3D local descriptors, 3D global descriptors encapsulate comprehensive and global information of the entire scene. Most handcrafted global descriptors describe places with global statistics of LIDAR scans. 
\cite{rohling2015fast} proposes a fast method of describing places through histograms of point elevation. 
DELIGHT \cite{palmer1998delight} designs a novel global descriptor by leveraging intensity information of LiDAR data. 
\cite{cao2018robust} converts 3D points to 2D images and then extracts ORB features from these images for scene correspondence.
With breakthroughs of learning based image retrieval methods, deep learning on 3D global descriptors for retrieval tasks has drawn growing attention. 
PointNetVlad \cite{angelina2018pointnetvlad} first tackles 3D place recognition in an end-to-end way, which combines PointNet \cite{qi2017pointnet} and NetVlad \cite{arandjelovic2016netvlad} to extract global descriptors from 3D points. 
Following this, PCAN \cite{zhang2019pcan} proposes an attention mechanism for local features aggregation, discriminating local features that contribute positively. 
However, these two methods employ PointNet architecture for extracting local features, which does not particularly concern the local geometry. 
LPD-Net~\cite{liu2019lpd} extracts the local contextual relationships but relies on handcrafted features.
DH3D \cite{du2020dh3d} designs a deep hierarchical network to produce more discriminative descriptors. 
%But this method adopts a two-stage strategy, which first trains local descriptors and then rectified global descriptors. 
DAGC \cite{sun2020dagc} introduces a graph convolution module to encode local neighborhood information. 
However, it does not count the spatial relationship between local descriptors. 
Compared with previous studies, our network combines the strengthens of their designs,  facilitating discriminative and versatile global descriptors.
% \begin{figure*}
% \begin{center}
% \fbox{\rule{0pt}{2in} \rule{.9\linewidth}{0pt}}
% \end{center}
%   \caption{Example of a short caption, which should be centered.}
% \label{fig:short}
% \end{figure*}

%------------------------------------------------------------------------
\section{Problem Statement} \label{sec: problem statement}

Let \textit{${M_{ref}}$} be a pre-built reference map of 3D point clouds defined with respect to a fixed reference frame, which is divided into a set of submaps such that ${ M_{ref} = {\left \{ m_{i}: i = 1,..., M \right \}}}$. %Notably, the numbers of points in each submap are the same.
The submap coverages are kept approximately the same. Each submap is tagged with a UTM coordinate at its centroid using GPS/INS. Let \textit{${Q}$} be a query point cloud with the same coverage with respect to a submap in \textit{${M_{ref}}$}. 
We define the place recognition problem as retrieving a submap \textit{${m^{*}}$} from \textit{${M_{ref}}$} that is structurally closest to \textit{${Q}$}. 
Note that under this formulation, \textit{${Q}$} is not a subset of \textit{${M_{ref}}$}, since they are independently scanned at different times. 

To tackle this problem, we design a neural network to learn a function $f(\cdot )$ that embeds a local point cloud to a 3D global descriptor of pre-defined size. 
The goal is to find a submap ${m^{*}\in M_{ref}}$ such that the distance between global descriptors ${f(m^{*})}$ and ${f(Q)}$ is minimized:
\begin{equation} 
    m^{*}=\underset{m_{i}\in M_{ref}}{{\rm argmin}}\: d(f(Q), f(m_{i})),
\label{Eq:encoder}
\end{equation}
\noindent 
where $d(\cdot )$ is a distance metric (e.g., Euclidean distance).
In practical implementation, a global descriptor dictionary is built offline for all 3D submaps. 
When a query scan appears, the nearest submap is obtained efficiently by comparing the global descriptor extracted online from the query scan with stored global descriptors.

%-------------------------------------------------------------------------
\section{SOE-Net} \label{sec: SOE-net}
\begin{figure*}[ht!]
\centering
\includegraphics[width=16cm]{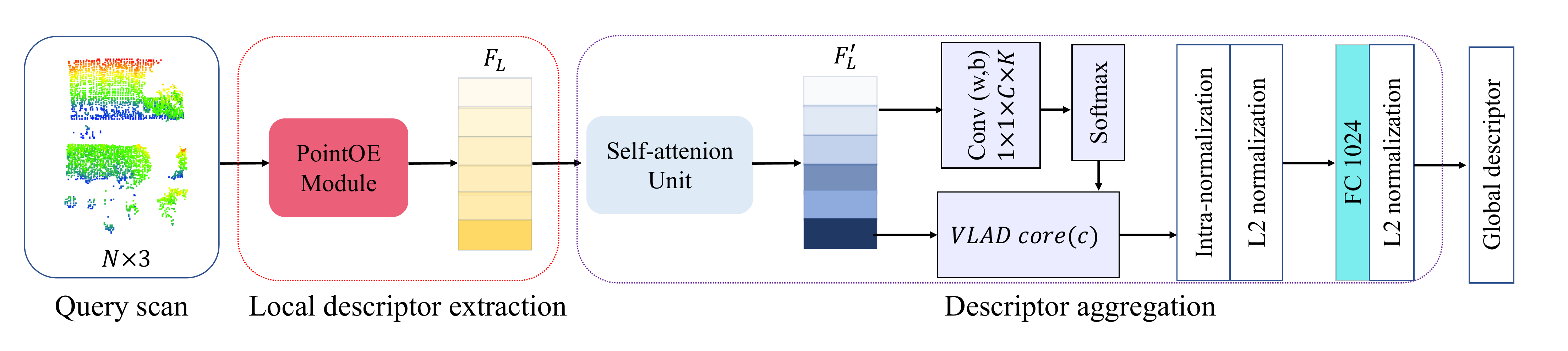}    
\caption{Overview of the SOE-Net architecture. The network takes a query scan with $N$ points as input and employs the PointOE module to extract point-wise local descriptors $F_{L}$. During descriptor aggregation, a self-attention unit is applied on the local descriptors and followed by the NetVLAD layer. Finally, a fully connected (FC) layer is adopted to compress the output descriptor vector, follow by the L2 normalization to produce a global descriptor.} 
\label{fig:network} 
% \vspace{-0.5cm}
\end{figure*}

Fig.~\ref{fig:network} shows the overall network architecture of our SOE-Net, where the local descriptor extraction part produces local descriptors from the 3D query scan, and the descriptor aggregation part aims to generate a distinct global descriptor. 

Given the input as a query point cloud with coordinates denoted as $Q=\left \{ p_{1},\cdots , p_{N} \right \}\in \mathbb{R}^{N \times 3} $, we first use the designed PointOE module to extract point-wise local descriptors. 
Unlike previous studies, it extracts relevant local information from eight directions to enhance point-wise feature representation, with details described in Section~\ref{sec: PointOE}. 
Then we propose a self-attention unit in the descriptor aggregation part to encode the spatial relationship among point-wise local descriptors, which is explained in Section~\ref{sec: self-attetnion}. 
Afterwards, the NetVLAD layer is adopted to fuse enhanced local descriptors in Section~\ref{sec: NetVLAD layer}. 
The training strategy with the proposed HPHN quadruplet loss is presented in Section~\ref{sec: loss function}.

\subsection{Local descriptor extraction}\label{sec: local feature}

\subsubsection{PointOE Module}\label{sec: PointOE}
\begin{figure}[ht!]
\centering
\includegraphics[width=8cm]{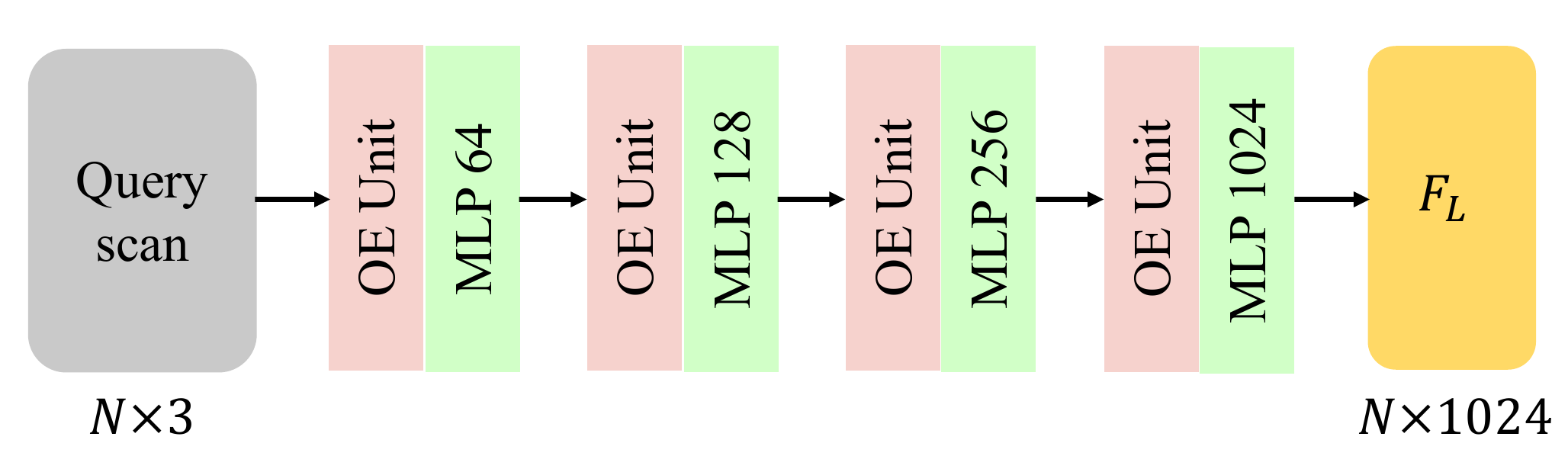}    
\caption{PointOE module. The input point cloud passes through a series of OE units and MLPs, local descriptors $F_{L}$ are generated as output. } 
\label{fig:pointoe} 
\end{figure}

\begin{figure}[ht!]
\centering
\includegraphics[width=8cm]{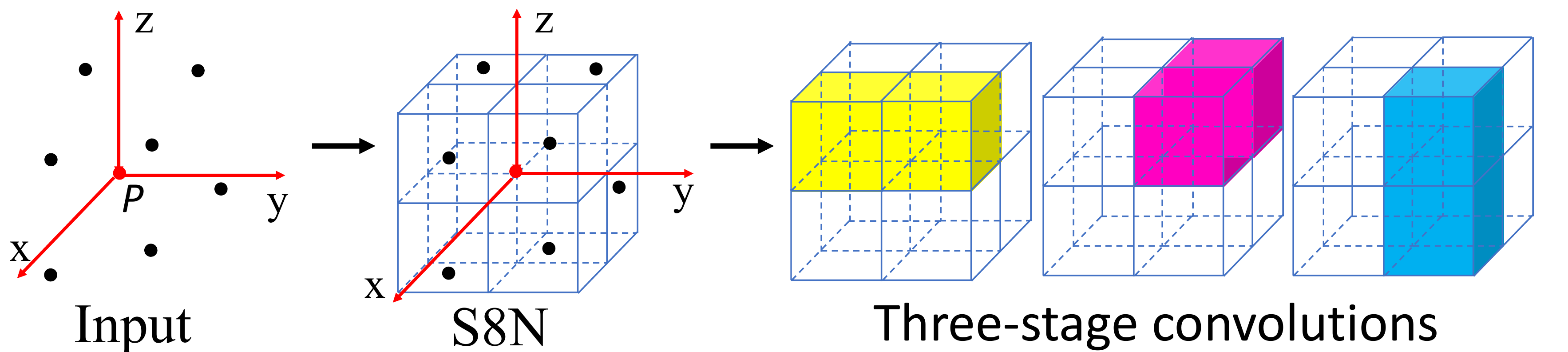}    
\caption{Illustration of OE unit.} 
\label{fig:oe} 
\end{figure}

The successes of many non-learning based image retrieval methods are owing to the design of great local image descriptors (e.g., SIFT \cite{lowe2004distinctive}). 
Orientation-encoding (OE) is one of SIFT's most shining highlights, which is also considered to benefit 3D feature description. 
Inspired by PointSIFT \cite{jiang2018pointsift}, we introduce the OE unit to the proposed SOE-Net. Specifically, we integrate it into PointNet to improve the point-wise feature representation ability. Fig.~\ref{fig:pointoe} shows the detailed architecture of the PointOE module. 
To the best of our knowledge, no prior work has explored it for large-scale place recognition and its effectiveness for retrieval has not been verified.

The inputs to our PointOE module are the 3D coordinates of $N$ points. Following \cite{qi2017pointnet}, multi-layer perceptrons (MLP) are adapted to encode the input 3D coordinates into features of $ {[64, 128, 256, 1024]} $ dimensions. 
We insert the OE unit in front of each MLP to improve the representation ability. 
Local descriptors $F_{L}$ are generated from this module.

{ \bf Orientation-encoding Unit.} Consider a $N \times C$ matrix as an input which describes a point cloud of size $N$ with a $C$-dimensional feature for each point, OE unit will output a feature map with the same dimension $N \times C$. 
Every point is assigned to a  new $C$-dimensional feature, which integrates the local information from eight orientations. 
As shown in Fig.~\ref{fig:oe}, the OE unit first adopts the Stacked 8-neighborhood Search (S8N) to find the nearest neighbors for point $P$ in each of the eight octants \cite{jiang2018pointsift}. 
% As pointed out by \cite{jiang2018pointsift}, ball query search in PointNet++ is for global nearest neighbors. Thus, it can cause the all chosen nearest points from the same direction, which is less informative than searching from 8 directions. 
Furthermore, we extract features using three-stage convolutions from those neighbors, which lie in a $2\times 2\times 2$ cube along the $x-,y-,z-$ axis. 
Formally, these three-stage convolutions are defined as:
\begin{equation}
\begin{aligned}
    OE_{x} &= {\rm ReLU}({\rm Conv}(w_{x},V,b_{x})), \\
    OE_{xy} &= {\rm ReLU}({\rm Conv}(w_{y},OE_{x},b_{y})), \\
    OE_{xyz} &= {\rm ReLU}({\rm Conv}(w_{z},OE_{xy},b_{z})),
\end{aligned}   
    \label{Eq:attention_map}
\end{equation}
\noindent
where $V\in \mathbb{R}^{2\times 2\times 2\times C}$ are the feature vectors of neighboring points. $w_{x}\in \mathbb{R}^{2\times 1\times 1\times C}, w_{y}\in \mathbb{R}^{1\times 2\times 1\times C}$ and $w_{z}\in \mathbb{R}^{1\times 1\times 2\times C}$ are weights of the three-stage convolutions, $b_{x}, b_{y}, b_{z}$ are the biases of convolution operators. 
By this way, the OE unit captures the local geometric structure from eight spatial orientations.

\subsection{Feature Aggregation} \label{feature aggregation}

\subsubsection{Self-attention Unit}\label{sec: self-attetnion}
\begin{figure}[ht!]
\centering
\includegraphics[width=8cm]{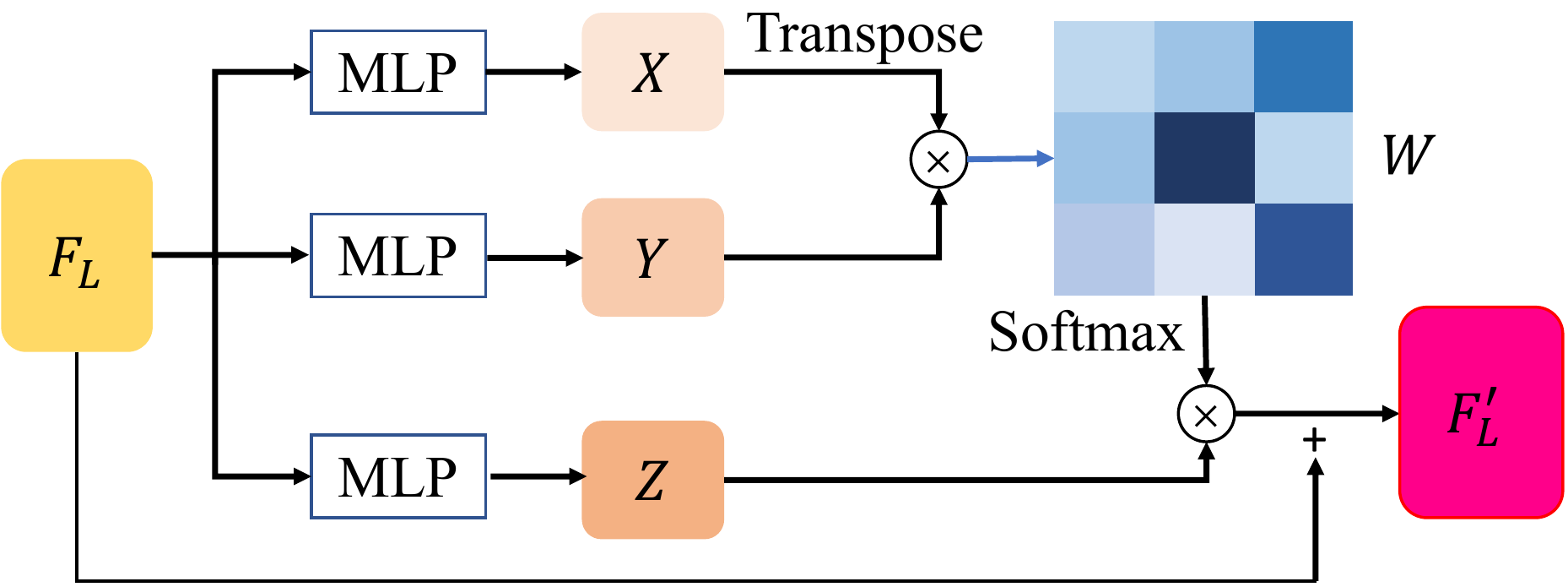}    
\caption{Illustration of the self-attention unit.} 
\label{fig: self_attetnion} 
\end{figure}

To introduce long-range context dependencies after extracting local descriptors, we design a self-attention unit \cite{zhang2019self} before fusing them into the NetVLAD layer. 
The self-attention unit can encode meaningful spatial relationships between local descriptors. 
Fig.~\ref{fig: self_attetnion} presents its architecture. 
Given local descriptors $F_{L}\in \mathbb{R}^{N \times C}$, where $N$ is the number of points and $C$ is the number of channels, we feed $F_{L}$ into two MLPs respectively and generate the new feature maps $X\in \mathbb{R}^{N \times C}$, $Y\in \mathbb{R}^{N \times C}$. 
Then the attention map $W$ is calculated, defined as follows:
\begin{equation}
    W_{j,i}=\frac{{\rm exp}(Y_{j}\cdot X_{i}^{T})}{\sum_{i,j=1}^{N}{\rm exp}(Y_{j} \cdot X_{i}^{T})},
    \label{Eq:attention_map}
\end{equation}
\noindent
where $W_{j,i}$ indicates that the $i^{th}$ local descriptor impacts on $j^{th}$ local descriptor, with the shape of $N \times N$. 
Here, it deems as the component that learns the long-range dependency relationship among point-wise local descriptors. 
More important local descriptors will contribute more to the representation of the target global descriptor. 
On the other hand, $F_{L}$ is fed into another MLP to output a new feature map $Z\in \mathbb{R}^{N \times C}$. 
Afterwards, we multiply it with the transpose of $W$ to generate the result $A^{P}\in \mathbb{R}^{N \times C}$. Finally, we add a scale parameter $\alpha$ on it and add  back $F_{L}$, which can be defined as follows:
\begin{equation}
    F_{L}^{'} = \mu A^{p} + F_{L}  
      = \mu W^{T}Z + F_{L},
    \label{Eq:attention_map}
\end{equation}
\noindent
where $\mu$ is initialized as zero and gradually assigned more weights with the progress of learning. 
The final output has a global context view compared with the original local descriptors. 
This enhances feature integration by combining geometrical and contextual information. 

\subsubsection{NetVLAD Layer} \label{sec: NetVLAD layer}
In this module, we aim to aggregate the local descriptors to a discriminative and compact global one. 
Following the configuration in \cite{angelina2018pointnetvlad}, we adopt a NetVLAD layer to fuse features. 
The NetVLAD layer learns $K$ visual words, denoted as $\left \{v_{1},\cdots , v_{K}|v_{k}\in \mathbb{R}^{C} \right \}$, and generates a $ ( C\times K )$-dimensional VLAD descriptor $F_{VLAD}=\left \{ F_{vlad}^{1},\cdots ,F_{vlad}^{K} \right \}$. 
However, the VLAD descriptor is time-consuming for nearest neighbor search, thus we apply a fully connected layer to generate a more compact global descriptor with an $L_{2}$ normalization.

\subsection{Loss function} \label{sec: loss function}

% \cite{xiao2017margin}
% \cite{chen2017beyond}
% \cite{schroff2015facenet}

Before going to the details of our proposed HPHN-quadruplet loss, we give a short review on the quadruplet loss~\cite{chen2017beyond} and its improvement. To compute the quadruplet loss, each batch of the training data includes $T$ quadruplets. Each quadruplet is denoted as $\Gamma_q = (\delta _{a}, \delta _{p}, \delta _{n}, \delta_{n}^{*} )$, where $\delta _{a}$ is a anchor point cloud, $\delta _{p}$ a positive point cloud (structurally similar to the query), $\delta _{n}$ a negative point cloud (structurally dissimilar to the query), $ \delta_{n}^{*} $ a randomly sampled point cloud that is different with $ \delta _{a}, \delta _{p}, \delta _{n}$. 
The quadruplet loss is formulated as:
\begin{equation}
\begin{split}
    L_{q}&\!=\!\frac{1}{T} \sum_{}^{T} \left[||f(\delta_{a})\!-\!f(\delta_{p})||_{2}^{2}\!-\!||f(\delta_{a})\!-\!f(\delta_{n})||_{2}^{2}\!+\!\alpha \right ]_{+} \\ 
    & \!+ \! \frac{1}{T} \sum_{}^{T} \left [||f(\delta_{a})\!-\!f(\delta_{p})||_{2}^{2}\!-\!||f(\delta_{n}^{*})\!-\!f(\delta_{n})||_{2}^{2}\!+\!\beta \right ]_{+},
\end{split}   
    \label{Eq:quadruplet}
\end{equation}
where $\left [  \cdots \right ]_{+}$ denotes the hinge loss, $\alpha$ and $\beta$ are the constant margins. 
The first term is a triplet loss which focuses on maximizing the feature distance between the anchor point cloud and the negative point cloud.
% in $\left \{ \delta _{n} \right \}$. 
The second term focuses on maximizing the feature distance between the negative point cloud and the additional point cloud $ \delta_{n}^{*} $.

To make the positive and negative samples in the quadruplet more effective, the quadruplet loss is extended to the lazy quadruplet loss~\cite{angelina2018pointnetvlad} by introducing hard sample mining. The quadruplets now become $\Gamma_{lq} = (\delta _{a}, \left \{ \delta _{p} \right \}, \left \{ \delta _{n} \right \}, \delta_{n}^{*} )$, where $\left \{ \delta _{p} \right \} $ is a collection of $\phi$ positive point clouds and $\left \{ \delta _{n} \right \} $ is a collection of $\varphi$ negative point clouds. The loss is modified accordingly to
% \begin{small}
\begin{equation}
\begin{split}
    L_{lq}&\!=\!\max_{\mathclap{\mbox{\tiny$\begin{array}{c}i=1...\phi \\ j=1...\varphi \end{array}$}}}(\left [||f(\delta_{a})\!-\!f(\delta_{p}^{i})||_{2}^{2}\!-\!||f(\delta_{a})\!-\!f(\delta_{n}^{j})||_{2}^{2}\!+\!\alpha \right ]_{+}) \\ 
    & \!+ \!\max_{\mathclap{\mbox{\tiny$\begin{array}{c}i=1...\phi \\ j=1...\varphi \end{array}$}}}(\left [||f(\delta_{a})\!-\!f(\delta_{p}^{i})||_{2}^{2}\!-\!||f(\delta_{n}^{*})\!-\!f(\delta_{n}^{j})||_{2}^{2}\!+\!\beta \right ]_{+}).
\end{split}   
    \label{Eq:lazy quadruplet}
\end{equation}
% \end{small}

%  \cite{angelina2018pointnetvlad, zhang2019pcan, du2020dh3d, sun2020dagc}. 

% In practice, $ \delta_{n}^{*} $ is sampled according to its spatial location: $ \delta_{n}^{*} $ should be far away from  $ \delta _{a}, \left \{ \delta _{p} \right \}$ and $\left \{ \delta _{n} \right \}$. 
% This, however, does not guarantee that the global descriptors of $ \delta_{n}^{*} $ is dissimilar to those of $ \delta _{a}, \left \{ \delta _{p} \right \}, \left \{ \delta _{n} \right \}$. 
% If $ \delta_{n}^{*} $ is structurally similar to $ \delta _{a} $, the second term is essentially the same as the first one in Eq.~\ref{Eq:lazy quadruplet}, thus the lazy quadruplet loss degenerates to a lazy triplet loss. 
% As seen from the top of Fig~\ref{fig:loss}, the lazy quadruplet loss can only push the negative point cloud $ \delta_{n}^{j}$ away from the positive point cloud $\delta_{p}^{i}$ in this situation.

In practice, a common strategy is to set $\beta$ to be smaller than $\alpha$ (e.g., $\alpha=0.5, \beta=0.2$) to make the second term in Eq.~\ref{Eq:lazy quadruplet} a relatively weaker constraint. However, we find this practice is less justified, especially in the scenario of metric learning for large-scale place recognition.   
In this work, we propose the Hard Positive Hard Negative quadruplet loss (HPHN quadruplet), which unifies the margin selection for $\delta _{a}$ and $ \delta_{n}^{*} $, and meanwhile rely on the hardest positive and the hardest negative samples in the batch to compute the learning signal. 
% \begin{figure}[ht]
% \centering
% \includegraphics[width=8.3cm]{loss.pdf}    
% \caption{Illustration of the effects of lazy quadruplet loss and the proposed HPHN quadruplet loss, respectively.} 
% \label{fig:loss} 
% % \vspace{-0.25cm}
% \end{figure}
In our case, the hardest positive point cloud $\delta_{hp}$ is the least structurally similar to the anchor point cloud, which is defined as:
 \begin{equation}
    \delta_{hp}=\underset{\delta_{p}^{i}\in\left \{ \delta_{p} \right \}}{ \rm argmax}\, \, ||f(\delta_{a})-f(\delta_{p}^{i})||_{2}^{2},
    \label{Eq:HP}
\end{equation}
\noindent
% where $\delta_{hp}$ is the hardest positive point cloud in $\left \{ \delta_{p} \right \}$.
The hardest negative point cloud is the most structurally dissimilar to the anchor point cloud. 
Here, we first find the hard negative point cloud $\delta_{hn}$ in $\left \{ \delta_{n} \right \}$, which is defined as:
 \begin{equation}
    \delta_{hn}=\underset{\delta_{n}^{j}\in\left \{ \delta_{n} \right \}}{\rm argmin}\, \, ||f(\delta_{a})-f(\delta_{n}^{j})||_{2}^{2}.
    \label{Eq:HN}
\end{equation}
Additionaly, we consider the feature distance from $\delta_{n}^{*}$ to $\delta_{n}$:
 \begin{equation}
    \delta_{hn}^{'}=\underset{\delta_{n}^{j}\in\left \{ \delta_{n} \right \}}{\rm argmin}\, \, ||f(\delta_{n}^{*})-f(\delta_{n}^{j})||_{2}^{2}.
    \label{Eq:HN_2}
\end{equation}
Finally, we select one of them as the hardest negative training data, which has the minimum distance $d_{hn}$:
 \begin{equation}
    d_{hn}= \rm min(||f(\delta_{a})-f(\delta_{hn})||_{2}^{2},||f(\delta_{n}^{*})-f(\delta_{hn}^{'})||_{2}^{2}).
    \label{Eq:d_HN}
\end{equation}

In conclusion, the HPHN quadruplet loss can be formulated as:
 \begin{equation}
   L_{HPHN} = \left [ ||f(\delta_{a})-f(\delta_{hp})||_{2}^{2} - d_{hn} + \gamma \right ]_{+},
   \label{Eq:HPHN}
\end{equation}
where $\gamma$ is the unified margin. 
The first term in Eq.~\ref{Eq:HPHN} is the upper bound of the feature distance of all the positive point cloud pairs, and the second term is the lower bound of the feature distance of all the negative point cloud pairs in a batch. 

Although having a form similar to the triplet loss, our loss is still a quadruplet loss that is computed from the sampled quadruplet. 
Compared with the lazy quadruplet loss, the proposed HPHN quadruplet loss picks the harder term between Eq.~\ref{Eq:HN} and Eq.~\ref{Eq:HN_2}, instead of using both in the loss computation. Moreover, the same margin is used when either of the both is selected. 
Despite this simple modification, our experimental results in Section \ref{subsection: Ablation study} demonstrate that our HPHN quadruplet loss significantly outperforms the lazy quadruplet loss.

\subsection{Implements Details} 
The proposed network is implemented in the Tensorflow framework and trained on a single Nvidia Titan Xp GPU with 12G memory. The size of the input points is 4096. The margins  $\gamma$ for the HPHN quadruplet loss are set to 0.5. Similar to all previous methods, we set the number of clusters $K$ in the NetVLAD layer to 64.
In the training stage, we set the batch size to 1 in each training iteration. Adam optimizer is used in the models for epoch 20. Same as PCAN, we choose 2 positive point clouds and 9 negative point clouds (including 1 other negative point cloud) in caculating loss functions. The initial learning rate is set to 0.0005. It is decayed by 0.7 after every 200K steps.

\section{Experiments}
\subsection{Benchmark Datasets}
We evaluate the proposed method on benchmark datasets proposed in \cite{angelina2018pointnetvlad}. 
It includes four open-source datasets for different scenes: Oxford RobotCar \cite{maddern20171} dataset and three in-house datasets of a university sector (U.S.), a residential area (R.A.), and a business district (B.D.). 
These datasets are all collected by a LiDAR sensor mounted on a car that travels around the regions repeatedly at different times. Based on the LiDAR scans, a database of submaps is built and each submap is tagged with a UTM coordinate. 
To better learn geometric features, the non-informative ground planes of all reference submaps are removed. 
The size of each submap is downsampled to 4096 points. 
In training, point clouds are regarded as correct matches if they are at maximum 10 m apart and wrong matches if they are at least 50 m apart. 
In testing, we regard the retrieved point cloud as a correct match if the distance is within 25 m between the retrieved point cloud and the query scan. 
% In Oxford RobotCar dataset, it includes 21771 training submaps and 3030 testing submaps using 44 sets of full and partial runs.  The training and testing submaps are split at regular intervals of 10 m and 20 m, respectively. The in-house datasets (U.S., R.A., and B.D.) consist of 400, 320, 200 submaps for training and 80, 75, 200 submaps for testing, respectively. The regular intervals are set to 12.5 and 25 m for training and testing submaps.
Following the experimental settings in \cite{angelina2018pointnetvlad, zhang2019pcan,sun2020dagc}, we first train our method using only the Oxford RobotCar training dataset. 
We will henceforth refer to this as our \textit{Baseline Network}. 
To improve the generalizability of the network, we further add the U.S. and R.A. training datasets into training data as our \textit{Refinement Network}.

\subsection{Results}
 \subsubsection{Baseline Network}
We compare our baseline network with PointNetVLAD (PN\_VLAD) \cite{angelina2018pointnetvlad} as a baseline and the state-of-the-art methods PCAN \cite{zhang2019pcan}, LPD-Net \cite{liu2019lpd}, DH3D \cite{du2020dh3d}, and DAGC \cite{sun2020dagc}. 
For a fair comparison, we use the same evaluation metrics, including the Average Recall at Top $N$ and Average Recall at Top 1\%. 
The final global descriptors of all networks are 256-dim. 
Table \ref{table:baseline} shows the top 1\% recall of each network on the four datasets. 
We refer to the recall values reported in \cite{du2020dh3d, sun2020dagc, liu2019lpd, zhang2019pcan, angelina2018pointnetvlad}. 
The recall values of DH3D for U.S., R.A. and B.D. are not reported in \cite{du2020dh3d}.
% \vspace{-0.25cm}
\begin{figure}[ht!]
\centering
\includegraphics[width=8.3cm]{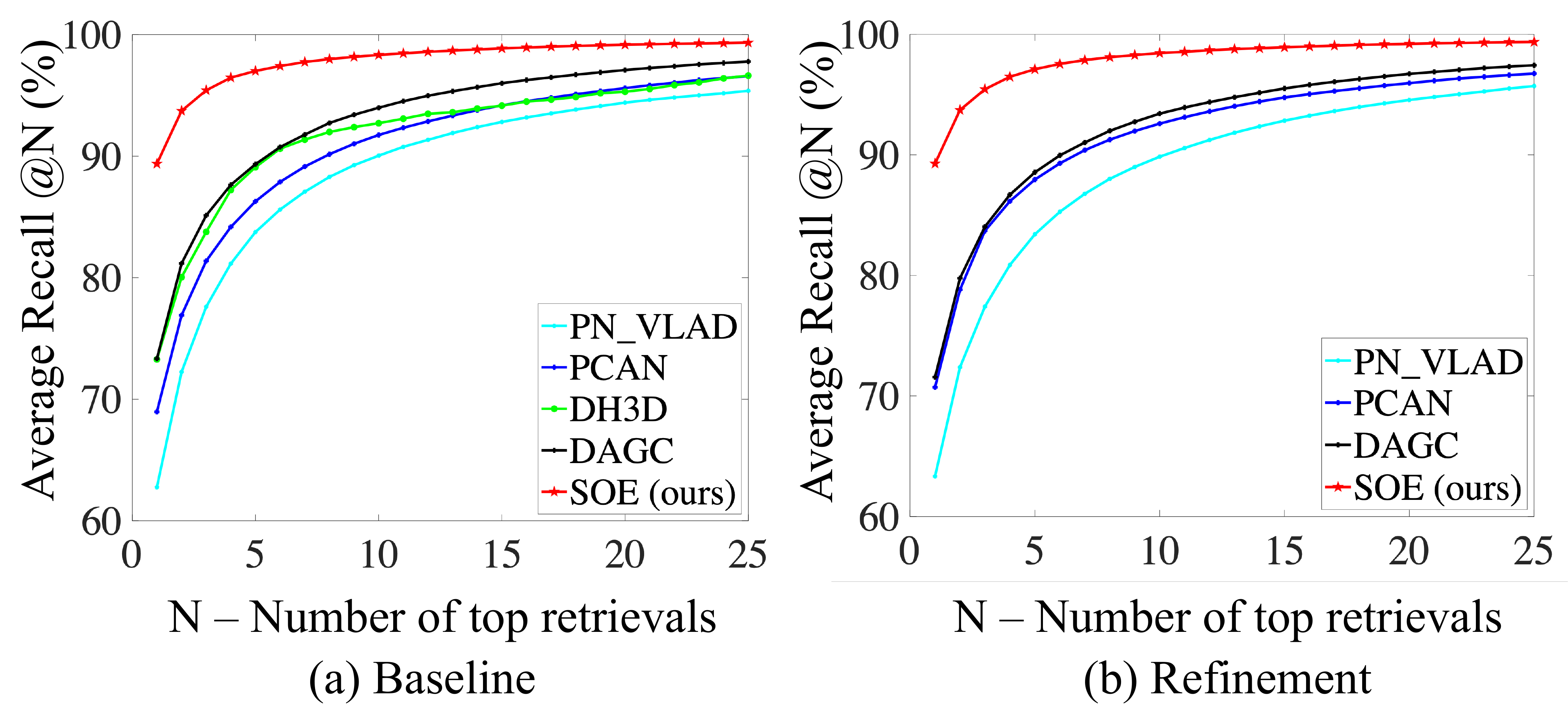}    
\caption{Average recall of SOE-Net tested on Oxford RobotCar. (a) shows the average recall when SOE-Net is only trained on Oxford RobotCar. (b) shows the average recall when SOE-Net is trained on Oxford RobotCar, U.S. and R.A. DH3D is not trained on this dataset in \cite{du2020dh3d}.} 
\label{fig: top25} 
\end{figure}

\vspace{-0.25cm}
\begin{table}[H]
\centering
\resizebox{0.47\textwidth}{!}{
\begin{tabular}{|c|c|c|c|c|c|c|} 
\hline
       & SOE             & DAGC  & DH3D  & LPD            & PCAN  & PN\_VLAD  \\ 
\hline
Oxford & \textbf{96.40}  & 87.49 & 84.26 & 94.92          & 83.81 & 80.31     \\ 
\hline
U.S.   & 93.17           & 83.49 & -     & \textbf{96.00} & 79.05 & 72.63     \\ 
\hline
R.A.   & \textbf{91.47}  & 75.68 & -     & 90.46          & 71.18 & 69.75     \\ 
\hline
B.D.   & 88.45           & 71.21 & -     & \textbf{89.14} & 66.82 & 65.30     \\
\hline
\end{tabular}}
\setlength{\abovecaptionskip}{0.cm}
\setlength{\belowcaptionskip}{0.5cm}
\caption{The average recall ($\%$) at top 1$\%$ for each network.}
\label{table:baseline}
\end{table}
\vspace{-0.15cm}

The results show that the proposed baseline network outperforms others significantly on Oxford RobotCar dataset.
The best performance on Oxford RobotCar reaches the recall of 96.40 $\%$ at top 1$\%$, exceeding the recall of current state-of-the-art method LPD-Net by 1.52 $\%$. Furthermore, SOE-Net achieves the recall of 93.17$\%$, 91.47$\%$, 88.45$\%$ on the unseen datasets respectively, which is similar or slightly weaker than LPD-Net. However, both of them improve the performance by a large margin compared with other methods. Notably, LPD-Net relies on ten handcrafted features, which has complex network architecture and high computational cost.
% Although the performance on the U.S. and B.D. is slightly weaker than LPD-Net, we think the performance on Oxford RobotCar dataset is more representative since the number of testing submaps on the U.S. and B.D. is too small.
Fig.~\ref{fig: top25}~(a) shows the recall curves of PointNetVLAD, PCAN, DAGC, and SOE-Net for the top 25 retrieval results. 
Notably, the recall at top 1 of SOE-Net reaches a recall of 89.37\%, indicating the proposed network effectively captures the task-relevant local information and generate more discriminative global descriptors.
More qualitative results are given in Section~\ref{sec: Results visualization}.

% \vspace{-0.25cm}
\subsection{Refinement Network} \label{sec: Refinement Network}
% \vspace{-0.25cm}
To improve the generalizability of the network on the unseen scenarios, \cite{zhang2019pcan, angelina2018pointnetvlad, sun2020dagc} further add U.S. and R.A. to the training data. 
We follow the same training sets to train our refinement network. 
As illustrated in Table \ref{table:refinement}, SOE-Net still significantly outperforms the state-of-the-art method DAGC on all datasets. 
By comparing Table \ref{table:baseline} and Table \ref{table:refinement}, it becomes clear that adding more data from different scenarios improves the performance of SOE-Net on the unseen dataset B.D.. 
In other words, given more publicly accessible datasets of real scans, SOE-Net has huge potential for LiDAR based localization. 
In Fig.~\ref{fig: top25}~(b) we plot the recall curves of the refinement network of PointNetVLAD, PCAN, DAGC, and SOE-Net for the top 25 retrieval results. 
It demonstrates that the global descriptors generated by SOE-Net are more discriminative and generalizable than all previously tested state-of-the-art methods.
% \vspace{-0.25cm}
\begin{table}[H]
\centering
\resizebox{0.47\textwidth}{!}{
\begin{tabular}{|c|c|c|c|c|c|c|} 
\hline
\multirow{2}{*}{} & \multicolumn{3}{c|}{~ ~Ave recall @1\%} & \multicolumn{3}{c|}{Ave recall @1~}  \\ 
\cline{2-7}
                  & SOE            & DAGC  & PCAN           & SOE            & DAGC  & PCAN        \\ 
\hline
Oxford            & \textbf{96.43}  & 87.78 & 86.40          & \textbf{89.28}  & 71.39 & 70.72       \\ 
\hline
U.S.              & \textbf{97.67}  & 94.29 & 94.07          & \textbf{91.75}  & 86.34 & 83.69       \\ 
\hline
R.A.              & \textbf{95.90}  & 93.36 & 92.27          & \textbf{90.19}  & 82.78 & 82.26       \\ 
\hline
B.D.              & \textbf{92.59}  & 88.51 & 87.00          & \textbf{88.96}  & 81.29 & 80.11       \\
\hline
\end{tabular}}
\setlength{\abovecaptionskip}{0.cm}
\setlength{\belowcaptionskip}{0.5cm}
\caption{Average recall ($\%$) at top 1$\%$ (@1$\%$) and top 1 (@1) for each of the models trained on Oxford RobotCar, U.S. and R.A..}
\label{table:refinement}
\end{table}
% \vspace{-0.25cm}

\subsection{Results visualization} \label{sec: Results visualization}
\begin{figure}[t]
\centering
\includegraphics[width=8.3cm]{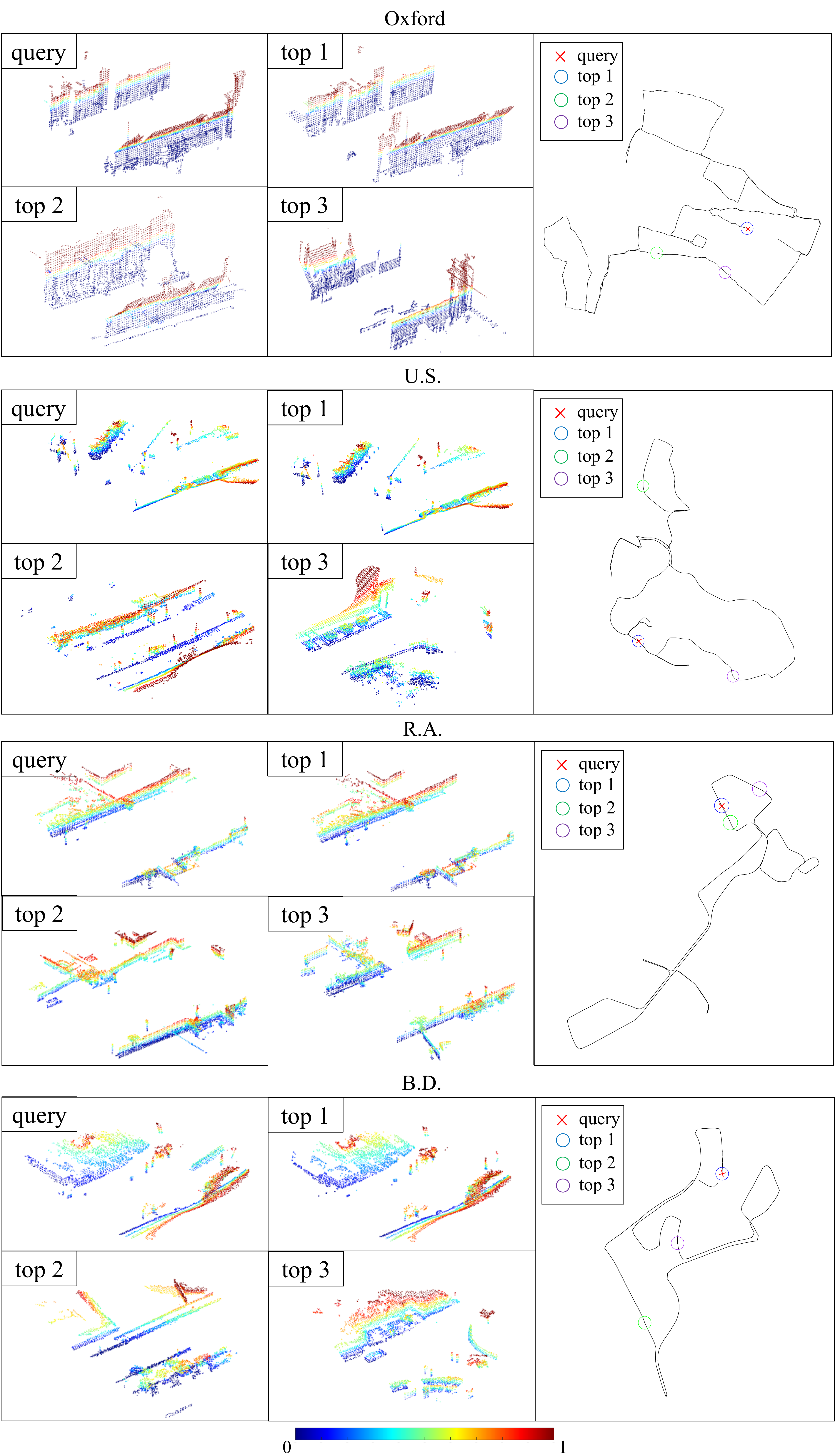}    
\caption{Visualizations of example retrieval results of SOE-Net on four benchmark datasets. For each retrieval, the query point cloud and the top 3 retrieved results are displayed. Locations of these point cloud is also indicated in the associated reference map. Colors in these point clouds represents heights above the ground.} 
\label{fig: results} 
\vspace{-0.25cm}
\end{figure}

% \vspace{-0.25cm}
In addition to quantitative results, we select and show qualitative results of some correctly retrieved matches in Fig.~\ref{fig: results}. 
A full traversal is chosen randomly as the reference map on four benchmark datasets, respectively. 
Then we choose four query point clouds from other randomly selected traversals on their respective datasets, with each representing one sample submap from individual testing areas. 
For each instance, the query point cloud and the top 3 retrieved matches are shown on the left. 
It becomes clear that the best match has a very similar scene as the query point cloud. 
Besides, we display the location of each point cloud in the reference map on the right. 
For each query, the location of the top 1 result (indicated by the blue circle) is correctly overlapped with the query location (represented by the red cross). 
It shows that the proposed network indeed has the ability to recognize places.

\section{Discussion} \label{section: Discussion}
% \vspace{-0.25cm}
\subsection{Ablation study}\label{subsection: Ablation study}
Ablation studies evaluate the effectiveness of different proposed components in our network, including both the PointOE module and self-attention unit. 
We also analyze the performance of the proposed HPHN quadruplet loss. 
All experiments are conducted on Oxford RobotCar.

% \vspace{-0.25cm}
{ \bf PointOE module and self-attention unit.} We test the effectiveness of the proposed PointOE module and the self-attention unit, using PointNetVLAD and PCAN as baselines (PN\_VLAD, PCAN).  
We first just integrate either PointOE module or self-attention unit into PointNetVLAD, referred as PN\_VLAD-OE and PN\_VLAD-S. 
We then combine both two components into PointNetVLAD, denoted as PN\_VLAD-SOE. Besides, we replace PointNet by PointNet++~\cite{qi2017pointnet++} in the local descriptor extraction stage, referred to as PN++\_VLAD.
All networks are trained with lazy quadruplet loss, with results shown in Table \ref{table:SOE}.
% \vspace{-0.25cm}

% \vspace{-0.25cm}
Comparing with PointNetVLAD, PN\_VLAD-S sees an improvement of 5.7\% and 10.27\% on the average recall at top 1\% and top 1, respectively. 
The performance of PN\_VLAD-S also exceeds the recall of PCAN by 2.9\% and 3.98\%, respectively, indicating the proposed self-attention unit is more effective than the attention strategy used in PCAN. 
This is due to the context information has a significant effect on aggregating local descriptors into a global one, and our self-attention unit can learn long-range spatial relationships between local descriptors.  
With the proposed PointOE module, our model brings significant improvements on the average recall by 11.19\% and 19.45\%, respectively,  when compared with PointNetVLAD. Besides, PointNet++ enhances PointNet features with a hierarchical encoding pipeline, but still does not explicitly encode orientation. The comparison with PN++\_VLAD demonstrates the superiority of OE for 3D descriptor learning for place recognition.
% This is consistent with our expectation that enhancing local descriptors is essential for generating a more discriminative global descriptor. 
Combining both modules can improve the performance by 12.40\% and 21.44\% on average recall, respectively. 
The ablation studies demonstrate the significant role of each module in SOE-Net.

\begin{table}[H]
\centering
\resizebox{0.47\textwidth}{!}{
\begin{tabular}{|c|c|c|} 
\hline
             & Ave recall @1\% & Ave recall @1  \\ 
\hline
PN\_VLAD     & 81.01           & 62.76          \\ 
\hline
PN++\_VLAD   & 89.10           & 76.23          \\ 
\hline
PCAN         & 83.81           & 69.05          \\ 
\hline
PN\_VALD-S   & 86.71           & 73.03          \\ 
\hline
PN\_VALD-OE  & 92.20           & 82.21          \\ 
\hline
PN\_VALD-SOE & \bf{93.41}   & \bf{84.20}           \\
\hline
\end{tabular}}
\setlength{\abovecaptionskip}{0.cm}
\setlength{\belowcaptionskip}{0.5cm}
\caption{Ablation studies of self-attetnion unit and PointOE module on Oxford RobotCar. The results show the average recall ($\%$) at top 1$\%$ (@1$\%$) and at top 1 (@1) for each model.}
\label{table:SOE}
\end{table}
% \vspace{-0.4cm}
%Removing any modules will decline the performance, which proves that each proposed module contributes. 
% \vspace{-0.25cm}

%\vspace{-0.25cm}
{\bf HPHN quadruplet loss.} To evaluate the proposed HPHN quadruplet loss, we compare the performance of the proposed SOE-Net trained with different losses. 
As shown in Table \ref{tab: loss}, the network performance is better when trained on the proposed HPHN quadruplet loss. 
The performance on Oxford RobotCar reaches 96.40\% recall at top 1\% and 89.47\% recall at top 1, exceeding the same model trained with the lazy quadruplet loss by 2.99\% and 5.17\%, respectively, demonstrating the superiority of the proposed HPHN quadruplet loss.
\begin{table}[H]
\centering
\resizebox{0.47\textwidth}{!}{
\begin{tabular}{|c|c|c|} 
\hline
              & \multicolumn{1}{l|}{Ave recall @1\%~} & Ave recall @1~  \\ 
\hline
Lazy quadruplet & 93.41                                  & 84.20            \\ 
\hline
HPHN quadruplet & \textbf{96.40}                         & \textbf{89.37}   \\
\hline
\end{tabular}}
\setlength{\abovecaptionskip}{0.cm}
\setlength{\belowcaptionskip}{0.5cm}
\caption{Results of the average recall ($\%$) at top 1$\%$ and at top 1 of SOE-Net trained with different losses on Oxford RobotCar.}
\label{tab: loss}
\end{table}
\vspace{-0.4cm}

\begin{table}[H]
\centering
\resizebox{0.47\textwidth}{!}{
\begin{tabular}{|c|c|c|c|c|c|c|} 
\hline
\multicolumn{1}{|l|}{} & \multicolumn{3}{c|}{SOE-Net}            & \multicolumn{3}{c|}{DAGC}       \\ 
\cline{2-7}
                       & D=128 & D=256          & D=512          & D=128 & D=256          & D=512  \\ 
\hline
Oxford                 & 95.30 & 96.40          & \textbf{96.70} & 84.43 & \textbf{87.49} & 85.72  \\ 
\hline
U.S.                   & 91.24 & 93.17          & \textbf{94.47} & 81.17 & \textbf{83.49} & 83.02  \\ 
\hline
R.A.                   & 90.53 & \textbf{91.47} & 91.00          & 72.39 & \textbf{75.68} & 74.46  \\ 
\hline
B.D.                   & 85.88 & 88.45          & \textbf{89.29} & 69.57 & \textbf{71.21} & 68.74  \\
\hline
\end{tabular}}
\setlength{\abovecaptionskip}{0.cm}
\setlength{\belowcaptionskip}{0.5cm}
\caption{Results of the average recall ($\%$) at top 1$\%$ of different global descriptor dimensions on Oxford RobotCar. D is the output dimension of global descriptors.}
\label{tab: dimension analysis}
\end{table}

\subsection{Output dimension analysis} \label{sec: dimension analysis}
In this section, we analyze the performance of the global descriptor with different output dimensions. 
The results of average recall at top 1\% for the global descriptor produced by SOE-Net and DAGC are shown in Table \ref{tab: dimension analysis}. 
We can draw two conclusions from this table:
(1) our method outperform DAGC, even if the generated global descriptor has a smaller dimension; %That demonstrates the superiority of the proposed network.
(2) when the output dimension decreases from 256 to 128, the performance of SOE-Net only declines by around 1\%-3\% on each benchmark. 
When the dimension expands to 512, the performance only changes by about 0.3\%-1\%. 
This implies the robustness of our method against different output dimensions. 
% \vspace{-0.25cm}

% \vspace{-0.25cm}

% \vspace{-0.25cm}
\subsection{Values of margin analysis} \label{sec: margin analysis}
% \vspace{-0.25cm}
In this section, we explore the network performance with different margins in the HPHN quadruplet loss using Oxford RobotCar. 
Table \ref{tab: margin analysis} shows results of average recall at top 1\% and top 1 with different margins for the SOE-Net architecture. 
Seen from the table, SOE-Net achieves the best performance with a margin value of 0.5. 
When the values expand to 0.7, the performance steadily degrades. 
This implies the distance between positive and negative pairs is sufficient with lower values of margin. 
On the other hand, when the value is set to 0.4, the performance decreases. 
So, we set the fixed value of margin as 0.5 in our network.
% \vspace{-0.25cm}
\begin{table}[H]
\centering
\setlength{\tabcolsep}{4.6mm}{
\begin{tabular}{|c|c|c|} 

\hline
Margin & \multicolumn{1}{l|}{Ave recall @1\%~} & Ave recall @1~  \\ 
\hline
0.4    & 95.87                                  & 88.84            \\ 
\hline
0.5    & \textbf{96.40}                         & \textbf{89.37}   \\ 
\hline
0.6    & 96.23                                  & 89.30            \\ 
\hline
0.7    & 95.63                                  & 88.46            \\
\hline
\end{tabular}}
\setlength{\abovecaptionskip}{0cm}
\setlength{\belowcaptionskip}{0.5cm}
\caption{Margin analysis in the HPHN quadruplet loss. We choose SOE-Net as a baseline and evaluate it on Oxford RobotCar.}
\label{tab: margin analysis}
\end{table}
% \vspace{0.25cm}

\section{Conclusion}
% \vspace{-0.25cm}
In this paper, we propose a novel end-to-end network SOE-Net for point cloud based retrieval. 
We design a PointOE module and a self-attention unit, using information from neighboring points and long-range context dependency to enhance the feature representation ability. 
In addition, we propose a novel HPHN quadruplet loss that achieves more discriminative and generalizable global descriptors. 
Experiments show that our SOE-Net improves the retrieval performance over state-of-the-art methods significantly. According to discussions on experimental results, especially ablation studies, we can discover that PointOE module contributes most to the performance of the SOE-Net.
There is also one notable limitation of the SOE-Net, which regards that the margin in the HPHN quadruplet loss needs to be set beforehand. 
In the future, we will explore adaptive margins that can better distinguish positive and negative pairs.

\vspace{0.5cm}
\noindent
{\bf Acknowledgement} We sincerely thank Max Hoedel for proofreading. This research was supported by the China Scholarship Council. This work was carried out within the frame of Leonhard Obermeyer Center (LOC) at Technische Universit{\"a}t M{\"u}nchen (TUM) [www.loc.tum.de].
{\small
\bibliographystyle{ieee_fullname}
\bibliography{egbib}
}

\end{document}